# Apples and Oranges? Assessing Image Quality over Content Recognition

Junyong You

Norwegian Researcher Centre, Bergen, Norway

Zheng Zhang

Hong Kong University of Science and Technology, Hong Kong, China

*Abstract*— Image recognition and quality assessment are two important viewing tasks, while potentially following different visual mechanisms. This paper investigates if the two tasks can be performed in a multitask learning manner. A sequential spatial-channel attention module is proposed to simulate the visual attention and contrast sensitivity mechanisms that are crucial for content recognition and quality assessment. Spatial attention is shared between content recognition and quality assessment, while channel attention is solely for quality assessment. Such attention module is integrated into Transformer to build a uniform model for the two viewing tasks. The experimental results have demonstrated that the proposed uniform model can achieve promising performance for both quality assessment and content recognition tasks.

*Keywords— content recognition, image quality assessment, multitask learning, Transformer, visual mechanisms*

## I. INTRODUCTION

Image content recognition has been the mainstream task in computer vision research. With the help of advanced deep learning models [1], the performance limitation of recognition has been pushed continuously, from representative CNNs, e.g., VGG, ResNet, InceptionNet, EfficientNet, to those Transformer based models, e.g., ViT [2], Swin [3]. The main advantage of deep networks for image recognition is that representative features can be extracted from spatial and channel dimensions gradually. On the other hand, image quality assessment (IQA), as a widely studied task, also benefits from deep learning. The mainstream approaches are quite similar to image recognition, i.e., using a sufficiently deep network to extract representative features followed by a dedicated head network for quality prediction. Most works on IQA attempt to directly follow the network architectures for the recognition task, e.g., using the same CNNs while replacing the head layer and retraining the networks on IQA databases [4]-[6]. As image recognition aims to classify an image into predefined categories, the head layer often employs multiple units with Softmax activation. Whereas, when predicting a single quality score for an image, the final layer of a deep network contains only one unit, and linear activation is used.

Selective attention plays an important role in viewing tasks. Attention describes that human perception can be more dominated by certain stimuli than others, determined by the stimuli or tasks [7][8]. Accordingly, computational attention models can be built in a bottom-up or a top-down approach [9]. As visual attention drives viewing behaviors, it has been represented in deep networks for CV tasks, e.g., different attention modules have been proposed to enhance the performance of CNNs for image recognition [10][11]. As a synthetical and revolutionary work, Transformer was first proposed for language modelling, e.g., machine translation, as certain words in the input text are more attentive to an output word/phrase [12]. Later, researchers have explored that Transformer can also reveal the most important feature representation for visual tasks. Consequently, many visual models are based on Transformer and its variations, e.g., ViT [2], Swin [3]. Naturally, it has attracted interest from the IQA community, and several IQA models inspired by Transformer have been proposed, e.g., TRIQ [13], MUSIQ [14].

Content recognition and IQA share certain similarity. For example, attention mechanism works in both tasks as aforementioned. Structural information is crucial for both content recognition [15] and quality assessment [16]. However, visual mechanisms can behave differently in the two tasks respectively, that differentiates the modelling approaches. Image down-rescaling is widely used in deep networks for content recognition. It is assumed that such rescaling does not affect the models to recognize the overall image content. However, image rescaling can potentially change the perceived image quality. For example, viewers often prefer high-dimension content on full-resolution screen than low-dimension ones. Thus, we intend to avoid rescaling images in IQA modelling. On the other hand, another visual mechanism, contrast sensitivity function (CSF), can play more important role in IQA than image recognition. This is because the masking effect introduced by CSF can conceal certain distortions. For example, same distortion occurred in a high-frequency area often causes different quality change than that in a low-frequency area [17]. Such effect is not significantly evident in content recognition. Thus, the differentiation between the two tasks should be considered when developing a uniform model.

Furthermore, multitask learning (MTL) has demonstrated promising applications in different areas, e.g., recommender systems [18], language modelling [19]. A typical example in visual modelling is object detection, where two tasks (object recognition and bounding box regression) are combined and trained simultaneously [20][21]. Due to the relationship between different tasks, the approach of parameter sharing (hard share or soft share and sharing degree) remains a challenge. To the best of our knowledge, no MTL studies for IQA and content recognition have been investigated, even though the authors of an IQA dataset, SPAQ [22], have superficially dipped into this issue.

In this work, we tackle the problem of representing IQA and content recognition in one uniform learnable model. An IQA-CR model driven by relevant visual mechanisms has been proposed, which is based on sharing essential image features while holding separate brunches for the two tasks to represent respective mechanisms, e.g., spatial attention and contrast sensitivity.

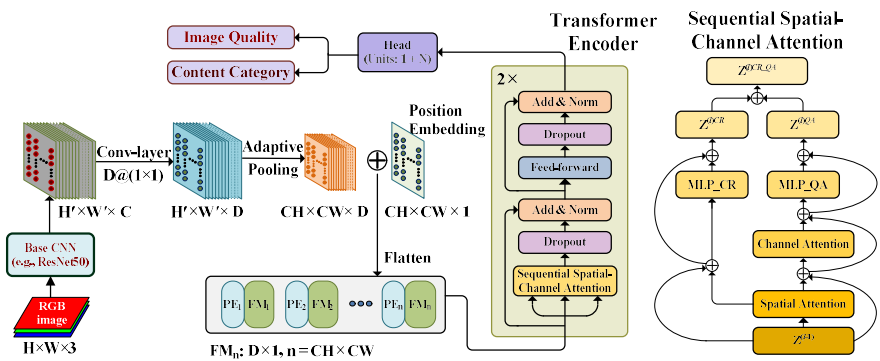

Fig.1. Architecture of the proposed IQA-CR model for IQA and content recognition, including transformer encoder and the flowchart of sequential spatial-channel attention. The number below each node indicates the output shape.

The model has been evaluated on two publicly available IQA databases with appending content categories. The experimental results clearly demonstrate promising performance on the two visual tasks simultaneously.

The remainder of the paper is organized as follows. Section II represents the relevant visual mechanisms briefly to support the joint IQA-CR modelling. Section III details the proposed IQA-CR model. Experiments are reported in Section IV. Section V draws concluding remarks finally.

## II. VISUAL MECHANISM IN CONTENT RECOGNITION AND IQA

It is intuitive that spatial attention plays an important role in both content recognition and quality assessment. Image content is dominated by certain salient objects [9], whilst viewers often pay more attention to objects of interest, e.g., those more distorted objects, in quality assessment [7][23][24]. Thus, applying spatial attention in both content recognition and IQA models can boost the performance. Many attention networks have been proposed that can represent unbalanced attention distribution over input images for content recognition or quality assessment [10][11][25][26]. A typical example is ViT [2], using the Transformer encoder on divided image patches to represent different importance of individual patches for the overall content category.

As explained earlier, perception on image distortion can be affected by the masking effect, which is determined by signal frequency and related visual mechanisms (e.g., foveation [7], attention guided eye movement [27]). The human visual system peaks its sensitivity at a critical frequency and drops beyond it [17]. Furthermore, several psychovisual experiments have demonstrated that visual sensitivity is affected by the attention mechanism as well. For example, [7] suggests that the critical frequency beyond which any contrast changes are imperceptible can be modified by attention distribution.

However, it is difficult to directly connect features to frequency domain by deep networks. In this work, we assume that the features in channel domain extracted by a CNN can simulate signal decomposition in frequency domain. An intuitive example is Sobel operator. Sobel operator is a convolution operation, which can distinguish high frequency components (e.g., edges) from low frequency components (e.g., plain areas) in an image. Our experiments will demonstrate that such assumption provides solid foundation to use deep networks for simulating the contrast sensitivity mechanism in IQA.

Subsequently, the contrast sensitivity mechanism can also be modelled by attention networks to represent unbalanced responses of an image signal to different frequencies. Furthermore, spatial and channel features can be combined in an attention network in IQA due to the mutual influence between spatial attention and contrast sensitivity mechanisms.

## III. IQA-CR MODEL

As explained in Section II, attention networks can be used to represent spatial attention and contrast sensitivity mechanisms, in which the former is essential for content recognition and IQA is affected by both two mechanisms. Thus, a uniform model for IQA and content recognition (IQA-CR) can be built following the above idea. Fig. 1 illustrates the model architecture.

Considering that channel features should be used to simulate frequency domain, a convolution network is used to extract spatial and channel features. This is different from other Transformer based models for content recognition (e.g., ViT) based on divided image patches. Another advantage of using convolution networks is that a convolution layer is independent of input size, only determined by its unit number, kernel size, stride, etc. As image rescaling potentially changes the perceived quality, using a convolution network can avoid the rescaling approach. In this work, any popular CNN can be used as the base network, e.g., ResNet, EfficientNet. The purpose of using CNN is to extract image features in spatial and channel domains. In addition, such approach is used to share low-level features for the content recognition and quality assessment tasks. We expect that such hard parameter sharing can represent certain similarity between the two tasks, whilst promoting the advantages of MTL.

A bottleneck convolution (kernel number $D$ and kernel size 1×1) is performed on the features extracted by the base CNN to reduce the channel dimension. Such approach has also been widely used in other attention networks, e.g., SEnet [28]. The bottleneck layer produces a feature map ($FM$) with shape of [$H'$, $W'$, $D$], where the spatial resolution $H'$ and $W'$ are determined by the original image size and the used base CNN. As image rescaling is not used before feeding an image into the base CNN, adaptively spatial pooling is performed on $FM$ to unify the feature map to a constant shape [$CH$, $CW$, $D$] for images with different resolutions. We can assume that each point on $FM$ conveys spatial and channel information from a patch in the original image, that will be fed into a Transformer encoder.

We can assume that swapping the spatial positions of image patches are crucial for content recognition and quality perception. Thus, spatial position embedding (*PE*) on *FM* is performed, as roughly explained. Eq. (1).

$$Z^{(0)}_{(s,c)} = [FM_1 + PE_1; \cdots FM_j + PE_j; \cdots] \quad (1)$$
$$F_j \in R^{D \times 1}, PE_j \in R^{H' \times W' \times D}$$

**Query/key/value computation.** In each layer $l$ of the Transformer encoder, a query/key/value vector in the multi-head attention (MHA) block is computed on the position embedded features $Z^{(l-1)}_{(s,c)}$ encoded by the proceeding layer.

$$\begin{cases} q^{(l,a)}_{(s,c)} = W^{(l,a)}_Q \cdot LN[Z^{(l-1)}_{(s,c)}] \in R^{D_h} \\ k^{(l,a)}_{(s,c)} = W^{(l,a)}_K \cdot LN[Z^{(l-1)}_{(s,c)}] \in R^{D_h} \\ v^{(l,a)}_{(s,c)} = W^{(l,a)}_V \cdot LN[Z^{(l-1)}_{(s,c)}] \in R^{D_h} \end{cases} \quad (2)$$

where $s$ and $c$ denote the spatial and channel dimensions, $a = 1, 2, \cdots, A$ is an index of multi-attention heads, the latent dimensionality of each head is determined as $D_h = D/A$, and $LN$ indicates layer normalization.

**Encoding.** The encoded representation of $Z^{(l)}_{(s,c)}$ at layer $l$ can be computed by a weighted sum of the vectors of self-attention coefficients from each head:

$$s^{(l,a)}_{(s,c)} = a^{(l,a)}_{(s,c),(0,0)} \cdot v^{(l,a)}_{(0,0)} + \sum_{s'=1}^{S} \sum_{c'=1}^{C} a^{(l,a)}_{(s,c)(s',c')} \cdot v^{(l,a)}_{(s',c')} \quad (3)$$

Next, the vectors from all heads are concatenated and passed through a multi-layer perceptron (MLP) layer using a residual connection, as roughly explained in Eq. (4) and Eq. (5).

$$Z'^{(l)}_{(s,c)} = W \begin{bmatrix} s^{(l,1)}_{(s,c)} \\ \vdots \\ s^{(l,A)}_{(s,c)} \end{bmatrix} + Z^{(l-1)}_{(s,c)} \quad (4)$$

$$Z^{(l)}_{(s,c)} = MLP[LN(Z'^{(l)}_{(s,c)})] + Z'^{(l)}_{(s,c)} \quad (5)$$

A key issue in Transformer encoder is to compute self-attention weights for encoding. In this work, a sequential spatial-channel self-attention module is proposed, which avoids computing spatial and channel attention separately to reduce the quadratic complexity of Transformer regarding the input length. The sequential approach can significantly reduce the computational complexity from $O(CH \times CW \times D)$ for separate spatial and channel attention to $O(CH \times CW + D)$.

**Sequential spatial-channel self-attention.** The spatial attention can be first computed by comparing each point $(s, c)$ with all the points at different spatial locations in the same channel.

$$a^{(l,a)\text{spatial}}_{(s,c)} = SM\left(\frac{q^{(l,a)^T}_{(s,c)}}{\sqrt{D_h}} \cdot [k^{(l,a)}_{(s,c)} \cdot \{q^{(l,a)}_{(s',c)}\}_{s'=1,\ldots,S}]\right) \quad (6)$$

As only spatial attention mechanism is used for image content recognition, the encoding $Z'^{(l)\text{spatial}}_{(s,c)}$ derived from the spatial attention weights in Eq. (4) is passed to a MLP layer (*MLP_CR*) in the Transformer encoder to derive the encoding $Z^{(l)CR}_{(s,c)}$ for content recognition from the layer $l$. However, both spatial attention and contrast sensitivity mechanisms are important for quality assessment, the encoding $Z'^{(l)\text{spatial}}_{(s,c)}$ is fed back for channel attention by Eq. (7).

$$a^{(l,a)\text{channel}}_{(s,c)} = SM\left(\frac{q^{(l,a)^T}_{(s,c)}}{\sqrt{D_h}} \cdot [k^{(l,a)}_{(s,c)} \cdot \{q^{(l,a)}_{(s,c')}\}_{c'=1,\ldots,C}]\right) \quad (7)$$

The resulting vector $Z'^{(l)\text{channel}}_{(s,c)}$ passes through another MLP layer (*MLP_QA*) to derive the encoding $Z^{(l)QA}_{(s,c)}$ for quality assessment from the layer $l$. It is noted that the computation of encodings $Z^{(l)CR}_{(s,c)}$ and $Z^{(l)QA}_{(s,c)}$ are using the same method as in Eq. (5) while two different MLP layers (*MLP_CR*, *MLP_QA*) are used. In this way, the model weights representing same mechanism, i.e., spatial attention, can be shared (soft parameter sharing) between content recognition and quality assessment tasks, while the channel attention weights representing contrast sensitivity mechanism is solely for quality assessment task. Next, the sum $Z^{(l)CR\_QA}_{(s,c)} = Z^{(l)CR}_{(s,c)} + Z^{(l)QA}_{(s,c)}$ is fed into the next encoder layer.

Earlier studies of Transformer for classification tasks, e.g., BERT [29], ViT [2], an extra token is added in the beginning of the positionally embedded vector. Recent work tends to use the average of Transformer encoder output along the attended dimension, rather than adding the extra token, e.g., Swin [3], Reformer [30]. This work follows such approach. The summed encoding $Z^{(L)CR\_QA}_{(s,c)}$ from the last layer $L$ are first averaged along the attended dimension. Finally, a fully connected head layer is applied to the concatenated vector to perform quality assessment and content recognition. The unit number of the head layer is determined by 1 plus the number of content categories ($N$), in which the first output value predicts a single image quality score whilst the rest values are probabilities of content categories. Accordingly, MSE is chosen as the loss function for quality assessment, while focal loss [21] is used in content recognition as class imbalance occurs in our datasets.

A tricky issue in MTL is to define the loss function. In this study, following the approach in image object detection where object classification and bounding box prediction are both performed, the sum of focal loss and MSE after L1-smooth has been used as a single loss for training the uniform model. We have observed that it produces better performance than using the two losses separately. In fact, other advanced loss combination strategies have also been tested in our experiments, e.g., dynamic weight average [31], uncertainty weighting methods [32], while no significant performance gain was obtained.

## IV. EXPERIMENTS AND DISCUSSION

### A. Datasets

Two datasets were used in our experiments. SPAQ [22], the only one publicly available dataset, to the best of our knowledge, contains both ground-truth content recognition (nine categories) and quality assessment, and KonIQ-10k [4]. Swin-B pretrained on ImageNet-22K [3] has been applied on KonIQ-10k images to classify the images preliminarily. Manual validation and adjustment have then been performed to finally classify the contained images into ten categories: animal, artifact, cityscape, human, indoor scene, landscape, night scene, plant, vehicle, and others.

The two datasets were split into train, validation and test sets, respectively. To avoid the long tail issues, the images in each dataset were *randomly* split according to their complexity (defined by the spatial perceptual information [33]) and MOS values. The images were first roughly classified into two complexity levels: high and low, and then the images in each level were further divided into five subcategories based on their MOS values. Finally, we randomly chose 80% of the images in each subcategory as train set, 10% as validation set and the rest 10% as test set. Such random split was performed for ten times.

TABLE I. AVERAGE EVALUATION RESULTS OF IQA MODELS ON THE TEST SET IN INDIVIDUAL DATABASES

| Models | SPAQ test sets | | | KonIQ-10k test sets | | |
|---|---|---|---|---|---|---|
| | PLCC | SROCC | RMSE | PLCC | SROCC | RMSE |
| **IQA-CR** | 0.919 | 0.923 | 0.330 | 0.920 | 0.911 | 0.235 |
| IQA-CR-Q | 0.922 | **0.927** | **0.326** | **0.935** | **0.918** | **0.202** |
| TRIQ | 0.916 | 0.925 | 0.324 | 0.922 | 0.910 | 0.223 |
| MUSIQ | 0.920 | 0.918 | 0.339 | 0.925 | 0.913 | 0.216 |
| DeepBIQ | 0.858 | 0.861 | 0.389 | 0.873 | 0.864 | 0.284 |
| Koncept512 | 0.831 | 0.830 | 0.384 | 0.916 | 0.909 | 0.267 |
| CaHDC | 0.824 | 0.815 | 0.486 | 0.856 | 0.817 | 0.370 |
| AIHIQnet | **0.929** | 0.925 | 0.328 | 0.929 | 0.915 | 0.209 |

TABLE II. AVERAGE OF TOP-1 ACCURACY OF CONTENT RECOGNITION MODELS ON THE TEST SET IN INDIVIDUAL DATABASES

| Datasets | IQA-CR | IQA-CR-C | Resnet50 | ViT | Swin-B |
|---|---|---|---|---|---|
| SPAQ | 0.859 | 0.862 | 0.769 | 0.860 | **0.870** |
| KonIQ-10k | 0.818 | **0.825** | 0.736 | 0.819 | 0.823 |

TABLE III. AVERAGE EVALUATION RESULTS OF IQA-CR AND OTHER MODELS ON THE TEST SET IN INDIVIDUAL DATABASES IN MTL EXPERIMENT

| Models | SPAQ test sets | | | KonIQ-10k test sets | | |
|---|---|---|---|---|---|---|
| | PLCC | SROCC | Accu. | PLCC | SROCC | Accu. |
| **IQA-CR** | **0.919** | **0.923** | **0.859** | **0.920** | **0.911** | **0.818** |
| MT-S | 0.914 | 0.920 | 0.678 | 0.902 | 0.894 | 0.630 |
| TRIQ | 0.885 | 0.875 | 0.701 | 0.893 | 0.886 | 0.694 |
| MUSIQ | 0.843 | 0.856 | 0.785 | 0.884 | 0.890 | 0.743 |
| DeepBIQ | 0.854 | 0.848 | 0.804 | 0.863 | 0.827 | 0.765 |
| Koncept512 | 0.837 | 0.850 | 0.811 | 0.855 | 0.809 | 0.770 |
| CaHDC | 0.822 | 0.810 | 0.776 | 0.830 | 0.823 | 0.698 |
| hyperIQA | 0.846 | 0.859 | 0.684 | 0.829 | 0.843 | 0.684 |
| ResNet50 | 0.795 | 0.774 | 0.807 | 0.803 | 0.797 | 0.775 |
| ViT | 0.755 | 0.783 | 0.813 | 0.804 | 0.803 | 0.802 |
| Swin-B | 0.795 | 0.812 | 0.819 | 0.800 | 0.813 | 0.810 |

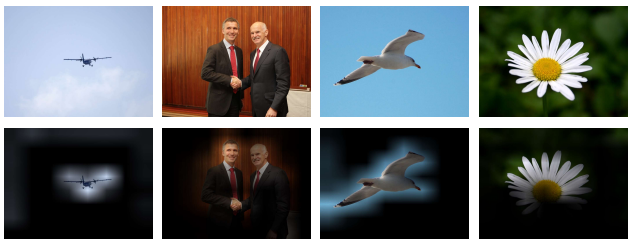

Fig. 2. Attention maps from the IQA-CA model.

### B. Experimental results

In our experiments, we have observed that a relatively simple architecture of IQA-CR model produces promising results, e.g., ResNet50 as base CNN and the Transformer encoder is set to [$L$=2, $D$=64, $H$=8, $d_{ff}$=128]. Larger models, e.g., more layers, large Transformer dimension, do not boost the performance significantly. We assume that this is due to the small scales of the used datasets cannot take full advantages of large models.

As a reference point, several state-of-the-art IQA and content recognition models have been compared in the experiments, including TRIQ [13], MUSIQ [14], DeepBIQ [5], Koncept512 [4], CaHDC [34], and AIHIQnet [25] for IQA, and Resnet50, ViT [2], and Swin-B [3] for content recognition. Besides, the proposed model can also be used for content recognition or quality assessment separately, given that the other brunch is removed. Thus, two modified models, namely IQA-CR-C for content recognition and IQA-CR-Q for quality assessment, were also tested in the experiments. Table I reports three evaluation criterions for IQA including Pearson correlation (*PLCC*), Spearman rank-order correlation (*SROCC*), and root-mean-squared-error (*RMSE*) between predicted MOS values and ground-truth, on the test sets in the datasets. Table II reports the top-1 accuracy for content recognition.

Furthermore, these content recognition and IQA models can also be modified to perform the two tasks simultaneously in a MTL manner. This is done by changing the unit number in the final head layer, as that in the proposed IQA-CR model. The results of these models in the MTL experiment are reported in Table III, in which MT-S is a MTL model proposed in the work of SPAQ dataset [22]. According to the evaluation results, the proposed IQA-CR model demonstrates comparable performance with other state-of-the-art models in separate tasks, and the modified models with removing one brunch even beat the comparable models. This confirms the feasibility and appropriateness of the visual mechanisms inspired modelling approach. More importantly, the results in Table III clearly show that IQA-CR model significantly outperforms other models in the MTL scenario by large margins. This demonstrates that the two viewing tasks: content recognition and quality assessment, indeed share certain similarity, and potentially confirms that human subjects can perform them simultaneously. However, the models dedicated to a single task (e.g., IQA or content recognition separately) with a simple modification for MTL do not perform satisfactorily, confirming that relevant mechanisms should be considered to build appropriate models for multiple tasks.

Finally, it is interesting to visualize attention maps showing what a deep model has learned during the training process. Fig. 2 shows attention maps from several images in KonIQ-10k dataset. As IQA-CR is a hybrid model consisting of CNN and Transformer encoder, an attention map is generated by fusing the gradient-weighted class activation map [35] from the base CNN and the attention weights derived from the self-attention layer in the Transformer encoder. The attention maps show that certain areas in an image are more important for visual tasks, and different areas have unbalanced contributions to the tasks.

### V. CONCLUSION

Inspired by two crucial visual mechanisms in viewing behaviors, selective attention and contrast sensitivity, this paper proposed a uniform model for IQA and content recognition. A sequential spatial-channel attention module is proposed to simulate the two mechanisms that can be appropriately integrated into Transformer encoder. Spatial attention is shared between content recognition and IQA, while channel attention is for IQA solely. The comparative results demonstrated that the proposed IQA-CR model shows outstanding performance in the content recognition and IQA tasks.